# Objection-Based Causal Networks

Adnan Y. Darwiche
Computer Science Department
Stanford University

## Abstract

This paper introduces the notion of objection-based causal networks, which resemble probabilistic causal networks except that they are quantified using objections. An objection is a logical sentence and denotes a condition under which a causal dependency does not exist. Objection-based causal networks enjoy almost all the properties that make probabilistic causal networks popular, with the added advantage that objections are, arguably, more intuitive than probabilities.

## 1 INTRODUCTION

Probabilistic causal networks (PCNs) [1, 11] have recently become very popular in many practical domains, such as medical diagnosis [8], vision [10], language understanding [7], and map learning [5]. A major difficulty in constructing these networks is providing prior and conditional probabilities that quantify root nodes and causal dependencies. Most people find these probabilities overly detailed and counterintuitive [6]. The work described in this paper attempts to overcome this difficulty by introducing the notion of objection-based causal networks (OCNs). Root nodes and causal dependencies in OCNs are quantified by providing objections rather than probabilities.

In PCNs, a causal dependency between nodes is quantified by providing a number in the interval $[0, 1]$. For example, the dependency from $A =$ "The grass is wet" to $B =$ "My shoes are wet" may be quantified by $p = .85$. The number $p$ is interpreted as a conditional probability of the effect $B$ given the cause $A$. Conditional probabilities are usually assessed by a domain expert or obtained from statistical data. However, most domain experts have difficulty assessing these probabilities and statistical data may not always be available [6, 1].

In OCNs, a causal dependency between nodes is quantified by providing a logical sentence under which the dependency does not exist. For example, the dependency from "The grass is wet" to "My shoes are wet" may be quantified by $C =$ "I did not step on the grass." The sentence $C$ is called an objection and plays the same role played by the number $p$ given above. The semantics of objections is based on the notions of objection-based states of belief and objection-based conditionalization [4], which are the counterparts of probabilistic states of belief and probabilistic conditionalization, respectively.

One could state at least three factors that make PCNs so popular. First, PCNs allow the representation of non-binary beliefs, which is not allowed by classical logic representations. Next, constructing a consistent PCN is much easier and more systematic than constructing a consistent classical logic theory. Finally, PCNs are based on probability theory, which supports many patterns of plausible reasoning [11, 12] that are not supported by classical logic.

OCNs enjoy all the above properties. Section 2 shows that an OCN can be interpreted as a state of belief that allows non-binary beliefs. Section 3 shows that constructing a consistent OCN is very similar to constructing a PCN. Finally, Section 4 shows how to compute a state of belief represented by an OCN, and Section 5 provides some concluding remarks.

## 2 OBJECTION-BASES STATES OF BELIEF

In probability calculus, we assess our confidence in a sentence by providing a number in the interval $[0, 1]$. If we have complete confidence in a sentence, we give it a probability of 1; otherwise, we give it a probability of less than one. Another way to assess our confidence in a sentence is by providing an *objection* to that sentence. For example, an objection to "Tweety is a bird implies Tweety flies" is "Tweety is wingless." This choice of assessing one's confidence leads to *objection calculus*, which underlies objection-based states of belief and causal networks.

Section 2.1 shows that an objection-based causal net-work can be interpreted as an objection-based state of belief. Sections 2.2, 2.3, and 2.4 discuss objection-based states of belief in more details.

### 2.1 TWO CLASSES OF CAUSAL NETWORKS

A quantified causal network is interpreted as a state of belief which maps a propositional language $\mathcal{L}$ into degrees of support.[1] The language $\mathcal{L}$ is formed from primitive propositions corresponding to nodes in the causal network and from logical connectives. For example, the network of Figure 1 has five nodes, $P_1, \ldots, P_5$, and the state of belief represented by that network has the sentence $P_1 \vee P_2 =$ "It rained or the sprinkler was on" in its domain.

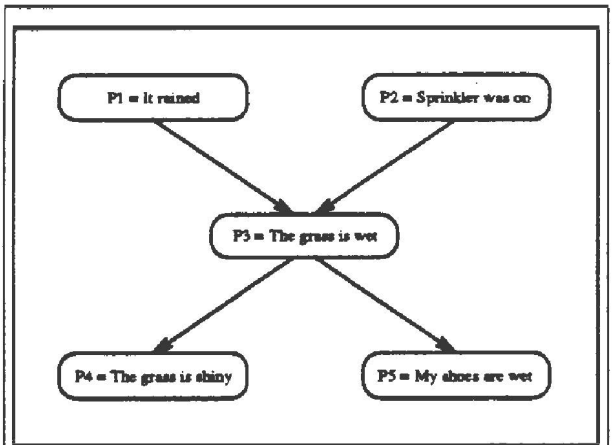

Figure 1: A causal network with five primitive propositions. Depending on how the network is quantified, it may become a probabilistic causal network or an objection-based causal network. The network is interpreted as a mapping from a language $\mathcal{L}$ into degrees of support (probabilities or objections). The language $\mathcal{L}$ is formed from primitive propositions $P_1, \ldots, P_5$ and logical connectives $\wedge, \neg, \vee$ and $\supset$.

If degrees of support are probabilities, then a state of belief and the causal network representing it are called probabilistic. Figure 1 and Table 2 constitute a PCN. On the other hand, if degrees of support are objections, then a state of belief and the causal network representing it are called objection-based. Figure 1 and Table 3 constitute an OCN.

From here on, I assume that objections are sentences in a propositional language $\mathcal{O}$. I also assume that primitive propositions of the languages $\mathcal{L}$ and $\mathcal{O}$ are disjoint.

---

[1] I assume that the support for a sentence does not determine that for its negation. It is common, though, to assume that the belief in a sentence determines that of its negation. I use the term "support" as opposed to "belief" to emphasize this difference.

### 2.2 OBJECTIONS

The notion of objection is central in objection-based states of belief. In this section, I discuss this notion and some related ones in more details.

If $\Phi : \mathcal{L} \longrightarrow \mathcal{O}$ is an objection-based state of belief, then $\Phi(A)$ is the objection of $\Phi$ to $A$. Here, $\Phi(A)$ is the reason why $\Phi$ has less than complete confidence in $A$. If $\Phi(A)$ is a tautology, we say that $\Phi$ *rejects* $A$ and *accepts* $\neg A$. And if $\Phi(A)$ is a contradictory sentence, we say that $\Phi$ has *no objection* to $A$.

If $\alpha$ is *the* objection of $\Phi$ to $A$, then any sentence entailing $\alpha$ is also *an* objection of $\Phi$ to $A$.

**Definition 1** *We say that a state of belief $\Phi$ objects to $A$ under $\alpha$ if and only if $\alpha \models \Phi(A)$.*

If the information available to a state $\Phi$ contradicts its objection to $A$, then under no further information will $\Phi$ object to $A$.

**Definition 2** *We say that a state of belief $\Phi$ admits $A$ under $\alpha$ if and only if $\alpha \models \neg\Phi(A)$.*

Let us consider an example. Suppose that the language $\mathcal{L}$ is defined over primitive propositions corresponding to the nodes of Figure 1, and suppose that the language $\mathcal{O}$ is defined over the following primitive propositions:

$O_1$ = "The grass is covered,"
$O_2$ = "The sprinkler is faulty,"
$O_3$ = "It is dark,"
$O_4$ = "I did not step on the grass,"
$O_5$ = "Something is abnormal."

If $\Phi(P_3 \supset P_5) = O_4$, then the objection of $\Phi$ to "The grass is wet implies my shoes are wet" is "I did not step on the grass." Moreover, $\Phi$ objects to "The grass is wet implies my shoes are wet" under "I did not step on the grass," and admits it under "I stepped on the grass."

### 2.3 CONSISTENCY

Probability theory imposes three consistency conditions on probabilistic states of belief. First, the probability of a tautology must be one. Second, the probability of a disjunction of logically disjoint sentences must be the sum of the probabilities of each of the disjuncts. Finally, equivalent sentences must have equal probabilities. An objection-based state of belief $\Phi$ has three corresponding conditions:

1. $\Phi(\text{true}) \equiv \text{false}$:[2] A tautology has a contradictory objection.

---

[2] I use **true** as a constant standing for any tautologous sentence and **false** as a constant standing for any contradictory sentence.

2. $\Phi(A \vee B) \equiv \Phi(A) \wedge \Phi(B)$: $\Phi$ objects to $A \vee B$ precisely when it objects to $A$ and to $B$.

3. $\Phi(A) \equiv \Phi(B)$ if $A \equiv B$: Equivalent sentences have equivalent objections.

### 2.4 OBJECTIONABILITY, BELIEF, AND IGNORANCE

I will conclude this section by discussing three important concepts: objectionability, belief, and ignorance. These concepts enrich the language used to describe states of belief.

In probability calculus we talk about a sentence being no more probable than another sentence. In objection calculus we talk about a sentence being no more objectionable than another.

**Definition 3**
*We say that $A$ is no more <u>objectionable</u> than $B$ in state $\Phi$ if and only if:*

$$\Phi(A) \models \Phi(B).$$

In probability calculus we say that a sentence is no more believed than another if and only if it is no more probable. In objection calculus, however, the relation between belief and objectionability is more involved.

**Definition 4** *We say that $A$ is no more <u>believed</u> than $B$ in state of belief $\Phi$ if and only if (a) $B$ is no more objectionable than $A$ in $\Phi$ and (b) $\neg A$ is no more objectionable than $\neg B$ in $\Phi$.*

The two conditions in Definition 4 may seem redundant, but this is not true. Table 1 provides a counterexample.

| $\mathcal{L}$ | $\mathcal{O}$ | $\mathcal{L}$ | $\mathcal{O}$ |
|---|---|---|---|
| bird $\wedge$ fly | $\neg$normal | bird | false |
| bird $\wedge \neg$fly | normal | $\neg$bird | false |
| $\neg$bird $\wedge$ fly | true | fly | $\neg$normal |
| $\neg$bird $\wedge \neg$fly | false | $\neg$fly | false |

Table 1: An objection-based state of belief. Here, $\neg$bird is no more objectionable than $\neg$fly, because the objection to $\neg$bird entails the objection to $\neg$fly (**false** $\models$ **false**). Note, however, that fly is more objectionable than bird, because the objection to bird strictly entails the objection to fly (**false** $\models \neg$**normal**).

In objection calculus, we can define the ignorance of a state of belief about a sentence as follows.

**Definition 5** *Let $\Phi : \mathcal{L} \to \mathcal{O}$ be an objection-based state of belief and let $A$ be a sentence in $\mathcal{L}$. The <u>ignorance</u> of $\Phi$ about $A$ is defined as follows:*

$$\Upsilon(A, \Phi) = \neg \Phi(A) \wedge \neg \Phi(\neg A).^3$$

---
[3] Note that $\Upsilon(A, \Phi)$ equals $\Upsilon(\neg A, \Phi)$.

Intuitively, $\Upsilon(A, \Phi)$ is the weakest sentence in $\mathcal{O}$ under which $\Phi$ admits both $A$ and $\neg A$.

At one extreme, $\Upsilon(A, \Phi)$ is a tautology. This means that $\Phi$ admits both $A$ and $\neg A$ under any sentence in $\mathcal{O}$, and we say that $\Phi$ is *maximally ignorant* about $A$. Note that maximal ignorance happens only when $\Phi$ has no objection to either $A$ or $\neg A$: $\Phi(A) \equiv \Phi(\neg A) \equiv$ **false**.

At the other extreme, $\Upsilon(A, \Phi)$ is a contradiction. This means that under no consistent sentence in $\mathcal{O}$ will $\Phi$ admit $A$ and $\neg A$. We say that $\Phi$ is *minimally ignorant* about $A$ in this case. Note that minimal ignorance happens only when $\Phi(A) \equiv \neg \Phi(\neg A)$.

**Definition 6** *We say that a state of belief $\Phi$ is no more <u>ignorant</u> about $A$ than about $B$ if and only if:*

$$\Upsilon(A, \Phi) \models \Upsilon(B, \Phi).$$

That is, whenever $\Phi$ admits $A$ and $\neg A$ it also admits $B$ and $\neg B$.

## 3 COMPONENTS OF A CAUSAL NETWORK

A quantified causal network has three components. One component is a directed graph such as the one depicted in Figure 1, which is referred to as an unquantified causal network. The topology of this graph contains information that partially defines a state of belief, and is the subject of Section 3.3. Another component of a causal network consists of prior degrees of support. These are either probabilities or objections attributed to root nodes of the network and are the subject of Section 3.1. The last component of a causal network consists of conditional degrees of support. These are either probabilities or objections that quantify causal dependencies in the network and are the subject of Section 3.2. The three components of a causal network fully define a state of belief.

### 3.1 PRIOR SUPPORTS

The first component of a causal network requires two degrees of support for each root node $P_i$. One of these degrees represents the support for $P_i$ and the other represents the support for $\neg P_i$. For example, the network of Figure 1 requires four prior degrees of support, which are attributed to each of $P_1 =$ "It rained," $\neg P_1 =$ "It did not rain," $P_2 =$ "The sprinkler was on," and $\neg P_2 =$ "The sprinkler was off."

In PCNs, prior supports are probabilities and must satisfy the following consistency condition: $\mathbf{P}(P_i) + \mathbf{P}(\neg P_i) = 1$. Tables 2 depicts prior probabilities for the network of Figure 1.

In OCNs, prior supports are objections and must satisfy the following consistency condition: $\Phi(P_i) \wedge \Phi(\neg P_i) \equiv$ **false**. This condition ensures that $P_i$ and $\neg P_i$ are not objected to simultaneously. Tables 3 depicts prior objections for the network of Figure 1.

| | | A | $P_3$ | $\neg P_3$ |
|---|---|---|---|---|
| $\mathbf{P}(P_1)$ | .5 | $\mathbf{P}(P_4 \mid A)$ | .75 | 0 |
| $\mathbf{P}(\neg P_1)$ | .5 | $\mathbf{P}(\neg P_4 \mid A)$ | .25 | 1 |
| $\mathbf{P}(P_2)$ | .5 | $\mathbf{P}(P_5 \mid A)$ | .9 | .5 |
| $\mathbf{P}(\neg P_2)$ | .5 | $\mathbf{P}(\neg P_5 \mid A)$ | .1 | .5 |

| A | $P_1 \wedge P_2$ | $P_1 \wedge \neg P_2$ | $\neg P_1 \wedge P_2$ | $\neg P_1 \wedge \neg P_2$ |
|---|---|---|---|---|
| $\mathbf{P}(P_3 \mid A)$ | .95 | .9 | .8 | 0 |
| $\mathbf{P}(\neg P_3 \mid A)$ | .05 | .1 | .2 | 1 |

Table 2: A probabilistic quantification of the causal network in Figure 1. The above probabilities and Figure 1 constitute a probabilistic causal network.

## 3.2 CONDITIONAL SUPPORTS

From here on, I refer to a set of propositions by their indices. For example, propositions $P_1, \ldots, P_k$ are referred to by $1, \ldots, k$. I also define a conjunction over a set of propositions $I$ to be a sentence in $\mathcal{L}$, denoted by $[\![I]\!]$, which has the form:

$$[\![I]\!] = \bigwedge_{i \in I} [\neg] P_i,$$

where $[\neg]$ means that the negation sign may or may not appear.[4] For example, there are four conjunctions over propositions 1, 2: $P_1 \wedge P_2$, $\neg P_1 \wedge P_2$, $P_1 \wedge \neg P_2$ and $\neg P_1 \wedge \neg P_2$.

The second component of a causal network requires $2^{n+1}$ degrees of support for each node $P_i$ with $n$ parents. Half of these degrees are supports for $P_i$ and the other half belongs to $\neg P_i$. Each of these supports is conditioned on a conjunction over the parents of $P_i$.

In PCNs, conditional degrees of support are conditional probabilities and must satisfy the following consistency condition: $\mathbf{P}(P_i \mid A) + \mathbf{P}(\neg P_i \mid A) = 1$. Table 2 depicts conditional probabilities for the network of Figure 1. For example, the conditional probability of $P_5 =$ "My shoes are wet" given $P_3 =$ "The grass is wet" is .9.

In OCNs, conditional degrees of support are conditional objections and must satisfy the following consistency condition: $\Phi_A(P_i) \wedge \Phi_A(\neg P_i) \equiv$ **false**. Here, $\Phi_A(P_i)$ and $\Phi_A(\neg P_i)$ are the conditional objections to $P_i$ and $\neg P_i$, respectively, given $A$. Table 3 depicts conditional objections for the network of Figure 1. For example, a conditional objection to $P_5 =$ "My shoes are wet" given $P_3 =$ "The grass is wet" is $O_4 =$ "I did not step on the grass."

We know the meaning of conditional probabilities, but what is the meaning of conditional objections?

---

[4] If $[\![I]\!]$ appears more than once in the same equation, then it refers to the same conjunction.

### 3.2.1 Objection-based conditionalization

A probabilistic state of belief $\mathbf{P}$ is changed using Bayes conditionalization, which states that the probability of $B$ after observing $A$ is $\mathbf{P}(B \wedge A)/\mathbf{P}(A)$.[5] Our goal in this section is to present a conditionalization rule for objection-based state of belief that is analogous to Bayes conditionalization, therefore, giving meaning to conditional objections.

Let $\Phi_A$ be the state of belief resulting from observing $A$ in the state $\Phi$. The least we should expect about the conditionalized state $\Phi_A$ is that it accepts $A$. Note, however, that there are many states of belief satisfying this property, and some of these states do not match our intuitions about belief change. We therefore need to impose more constraints on a conditionalized state of belief so that undesirable changes in belief are excluded.

I will now state a convention and an intuition about objection-based belief change. I will then present a conditionalization rule that is implied by the stated convention and intuition. First is the convention:

> An accepted sentence remains accepted after observing a non-rejected sentence.

By definition, a non-rejected sentence does not contradict any accepted sentence. The above convention says that none of the accepted sentences should be given up as a result of observing a non-rejected sentence.

Following is the intuition about changed objections:

> The objection to a sentence $B$, after observing a non-rejected sentence $A$, is the initial objection to $A \wedge B$ minus the initial objection to $A$.

The previous convention and intuition imply the following conditionalization rule:

$$\Phi_A(B) \equiv \begin{cases} \text{true}, & \text{if } \Phi(A \wedge B) \equiv \text{true}; \\ \Phi(A \wedge B) \wedge \neg \Phi(A), & \text{otherwise.} \end{cases} \quad (1)$$

Similar to Bayes conditionalization, objection-based conditionalization assumes that the observed sentence $A$ is not rejected by $\Phi$.

Objection-based conditionalization is an instance of abstract conditionalization [4], which supports patterns of plausible inference that make Bayes conditionalization popular [12].

### 3.2.2 Objection-based product rule

Equation 1 tells us how to change a state of belief upon recording an observation. Most often, however,

---

[5] Bayes conditionalization assumes that $A$ is not rejected by $\Phi$.

|  |  |  | $A$ | $P_3$ | $\neg P_3$ |
|---|---|---|---|---|---|
| $\Phi(P_1)$ | false |  | $\Phi_A(P_4)$ | $O_3$ | true |
| $\Phi(\neg P_1)$ | false |  | $\Phi_A(\neg P_4)$ | $\neg O_3 \wedge \neg O_5$ | false |
| $\Phi(P_2)$ | false |  | $\Phi_A(P_5)$ | $O_4$ | false |
| $\Phi(\neg P_2)$ | false |  | $\Phi_A(\neg P_5)$ | $\neg O_4 \wedge \neg O_5$ | false |

| $A$ | $P_1 \wedge P_2$ | $P_1 \wedge \neg P_2$ | $\neg P_1 \wedge P_2$ | $\neg P_1 \wedge \neg P_2$ |
|---|---|---|---|---|
| $\Phi_A(P_3)$ | $O_1$ | $O_1$ | $O_1 \vee O_2$ | true |
| $\Phi_A(\neg P_3)$ | $\neg O_1 \wedge \neg O_5$ | $\neg O_1 \wedge \neg O_5$ | $\neg O_1 \wedge \neg O_2 \wedge \neg O_5$ | false |

Table 3: An objection-based quantification of the causal network in Figure 1. The above objections and Figure 1 constitute an objection-based causal network.

we have information about a changed state of belief and we want to know more about the old state. In probability calculus, this is achieved by using an important result called the product rule. It states that

$$\mathbf{P}(A \wedge B) = \mathbf{P}(B|A) \times \mathbf{P}(A), \text{ if } \mathbf{P}(A) \neq 0.$$

The restriction on the equation above results from the inability to conditionalize a probabilistic state of belief on a sentence with zero probability.

A corresponding result in objection calculus is

$$\Phi(A \wedge B) \equiv \Phi_A(B) \vee \Phi(A), \text{ if } \Phi(A) \not\equiv \text{true and}$$

$$[\Phi_A(B) \equiv \text{true or } \Phi_A(B) \wedge \Phi(A) \equiv \text{false}]. \quad (2)$$

Note how logical disjunction is playing the role that is played by numeric multiplication in probability calculus. Note also that we have more restrictions on this rule than we had on the probabilistic one. Why? Well, the story goes as follows. Suppose that a domain expert told us that his state of belief is such that:

1. The objection to $A$ is $a$.
2. The objection to $B$ after observing $A$ is $b$, where $b$ is invalid and $a \wedge b$ is consistent.

The above statements are contradictory if the domain expert is using Equation 1 to update his state of belief. Specifically, according to Equation 1, $b$ must equal $\Phi(A \wedge B) \wedge \neg a$, which can happen only if $a \wedge b$ is inconsistent.

Contradictory statements of the form given above are a result of careless assessment of one's objections. An example will illustrate this point. We all know that having no wings is an objection to an animal being a bird:

$$\Phi(\text{bird}) = \text{wingless}.$$

We also know that abnormality is an objection to the flying of a bird:

$$\Phi_{\text{bird}}(\text{fly}) = \neg \text{normal}.$$

Although the two statements above seem plausible, they are in fact contradictory in the context of Equation 1. To see this, note that Equation 1 implies $\Phi_A(B) \models \neg \Phi(A)$. That is, the given conditional objection must entail the negation of the condition's objection. But abnormality does not entail having wings! The problem is that when assessing objections, we tend to forget the following important fact:

> An invalid conditional objection should entail the negation of the objection to the condition.

This, however, can be easily remedied if whenever a domain expert provides an invalid $b$ as the objection given $A$, we take that to mean $b$ conjoined with the negation of $A$'s objection: $b \wedge \neg \Phi(A)$.

### 3.3 IRRELEVANCE

The third component of a causal network is a directed graph such as the one depicted in Figure 1. The syntax of the graph does not depend on whether it is part of a PCN or an OCN, but its interpretation (the information it represents) does.

Informally, the directed graph of a causal network says the following: "Given a conjunction over the parents of a proposition $P_i$, information about the non-descendants of $P_i$ become irrelevant to the support for $P_i$." For example, considering the network of Figure 1, once we know that "The grass is wet," the information "It rained" does not change the support for "My shoes are wet."

Probability calculus formalizes irrelevance in terms of conditional probability. The statement "$A$ becomes irrelevant to $B$ once $C$ is known" is formalized as "The probability of $B$ given $C \wedge A$ equals the probability of $B$ given only $C$." Formally, irrelevance information represented by a PCN can be summarized by a single equation. If $D(i)$ contains the parents of proposition $P_i$, and $O(i)$ contains its non-descendants, then a PCN asserts that

$$\mathbf{P}([\![i]\!] \mid [\![D(i)]\!]) = \mathbf{P}([\![i]\!] \mid [\![D(i)]\!] \wedge [\![O(i)]\!]).$$

This says that $[\![O(i)]\!]$ becomes irrelevant to $[\![i]\!]$ once $[\![D(i)]\!]$ is accepted.

The objection-based formalization of irrelevance is similar to the probabilistic one. Specifically, $A$ becomes irrelevant to $B$ once $C$ is known if and only if (a) $B$'s objection given $C \wedge A$ is equivalent to $B$'s objection given $C$ and (b) $\neg B$'s objection given $C \wedge A$ is equivalent to $\neg B$'s objection given $C$.[6] Irrelevance information in an OCN is summarized by the following equation:

$$\Phi_{[D(i)]}([\![i]\!]) \equiv \Phi_{[D(i)] \wedge [O(i)]}([\![i]\!]). \qquad (3)$$

This says that the objection to $[\![i]\!]$ given the information $[\![D(i)]\!]$ is equivalent to the objection to $[\![i]\!]$ given the more detailed information $[\![D(i)]\!] \wedge [\![O(i)]\!]$.

## 4 FROM CAUSAL NETWORKS TO STATES OF BELIEF

Irrelevance information represented by the network of Figure 1 and the probabilities given in Table 2 constitute a complete definition of a probabilistic state of belief. This follows from a known result in the literature on causal networks (see [11] for example).

Similarly, irrelevance information represented by the network of Figure 1 and objections given in Table 3 constitute a complete definition of an objection-based state of belief. Proving this result is outside the scope of this paper — the interested reader is referred to [3] where the proof is given with respect to abstract causal networks that generalize PCNs and OCNs. However, I will show in the remaining of this section that the above claim is true for the OCN that was developed in Figure 1 and Table 3.

Consider the following equation, which is an instance of Equation 2:

$$\Phi([\![1,\ldots,5]\!]) \equiv \bigvee_{j=1}^{5} \Phi_{[\![1,\ldots,j-1]\!]}([\![j]\!]). \qquad (4)$$

Considering prior and conditional objections that are given in Tables 3, it is clear that we cannot evaluate Equation 4 because we do not have the objections needed. However, by using irrelevance information that is represented by the network of Figure 1, we can reduce Equation 4 to:

$\Phi([\![1,\ldots,5]\!]) \equiv$
$\Phi_{[3]}([\![5]\!]) \vee \Phi_{[3]}([\![4]\!]) \vee \Phi_{[1,2]}([\![3]\!]) \vee \Phi([\![2]\!]) \vee \Phi([\![1]\!]).$

All conditional and prior objections required by the above equation are given in Table 3. For example, $\Phi(P_5 \wedge P_4 \wedge P_3 \wedge \neg P_2 \wedge P_1)$ is equivalent to

$\Phi_{P_3}(P_5) \vee \Phi_{P_3}(P_4) \vee \Phi_{\neg P_2 \wedge P_1}(P_3) \vee \Phi(\neg P_2) \vee \Phi(P_1),$

which is also equivalent to $O_4 \vee O_3 \vee O_1$. That is, "Either I did not step on the grass or the grass is covered or it is dark" is the objection to "The sprinkler was off, but it rained, the grass is wet, shiny and my shoes are wet."

---

[6]The second part of the definition is not redundant. The reason is closely related to the example given in Section 2.4.

Probabilistically, $\mathbf{P}(P_5 \wedge P_4 \wedge P_3 \wedge \neg P_2 \wedge P_1)$ equals
$$\mathbf{P}(P_5|P_3) \times \mathbf{P}(P_4|P_3) \times \mathbf{P}(P_3|\neg P_2 \wedge P_1) \times$$
$$\mathbf{P}(\neg P_2) \times \mathbf{P}(P_1) = .151875. \qquad (5)$$

Since any sentence about the network of Figure 1 can be written as a disjunction of some instances of $[\![1,\ldots,5]\!]$, and since $\Phi(A \vee B) \equiv \Phi(A) \wedge \Phi(B)$, we can compute the objection to any sentence about the network of Figure 1. Figure 1 and Table 3 do specify an objection-based state of belief.

## 5 DISCUSSION

Objection-based causal networks resemble probabilistic causal networks in their structure and behavior. In an objection-based network, a dependency between nodes is quantified by providing logical sentences under which the dependency does not exist. Objection-based causal networks enjoy almost all properties that make probabilistic causal networks popular, with the added advantage that objections are, arguably, more intuitive than probabilities. Following are other advantages of objection-based causal networks over their probabilistic counterparts.

- *Incomparable supports.* Degrees of support in OCNs are only partially ordered while their probabilistic counterparts are totally ordered. In the network of Figure 1, probabilistic quantification has forced us to say that the causal dependency from "It rained" to "The grass is wet" is weaker than the one from "The grass is wet" to "My shoes are wet." No such commitment is enforced by objection-based quantification.

- *Intuitiveness.* Probably the major objection to PCNs is the interpretation of numbers that are used in quantifying causal dependencies. For example, what does it mean to say that the causal dependency from "The grass is wet" to "My shoes are wet" has strength .9? Or worse, what does the number .151875 that appears in Equation 5 mean [9]? This number does not relate in any intuitive way to the numbers used in quantifying the network. On the other hand, the strength of a causal dependency in an OCN has a very clear interpretation: it is a condition under which the causal dependency does not exist.

I end this discussion by observing that objection calculus and objection-based causal networks are closely related to clause management systems, which are well known in AI [2, 13].

## Acknowledgement

I have benefited from various discussions with my advisor Matt Ginsberg, Jinan Hussain and H. Scott

Roy read versions of this paper and provided valuable comments. Three anonymous reviewers have also provided valuable comments and suggestions. This work has been supported by the Air Force Office of Scientific Research under grant number 90-0363, by NSF under grant number IRI89-12188, and by DARPA/Rome Labs under grant number F30602-91-C-0036.